\documentclass{article}

\usepackage[letterpaper,margin=1in]{geometry}

\usepackage[utf8]{inputenc}
\usepackage[T1]{fontenc}
\usepackage{microtype}
\usepackage{natbib}
\usepackage{graphicx}
\usepackage{subcaption}
\usepackage{booktabs}
\usepackage{hyperref}
\hypersetup{hidelinks}
\usepackage{amsmath}
\usepackage{amssymb}
\usepackage{mathtools}
\usepackage{amsthm}
\usepackage{url}
\usepackage{multirow}
\usepackage{algorithm}
\usepackage{algorithmic}
\usepackage{float}
\usepackage{marvosym}
\usepackage[capitalize,noabbrev]{cleveref}
\usepackage{xcolor}
\usepackage{adjustbox}
\usepackage{placeins}
\usepackage{newunicodechar}
\newunicodechar{—}{---}
\newunicodechar{–}{--}
\newunicodechar{‐}{-}
\newunicodechar{’}{'}
\newunicodechar{“}{``}
\newunicodechar{”}{''}
\newunicodechar{…}{\ldots}

\theoremstyle{plain}

\theoremstyle{definition}

\theoremstyle{remark}

\title{Task-Level Natural Language Priors as Learning Signals for Low-Resource LLM Training}
\author{
    Jian Gao\textsuperscript{\rm 2},
    Xiao Zhang\textsuperscript{\rm 1},
    Xun Zhu\textsuperscript{\rm 1},
    Miao Li\textsuperscript{\rm 1 \Letter},
    Ji Wu\textsuperscript{\rm 1,2 \Letter}
    \\
    \textsuperscript{\rm 1}Department of Electronic Engineering, Tsinghua University \\
    \textsuperscript{\rm 2}College of AI, Tsinghua University \\
    \{gaojian21, xzhang19, zhu-x24\}@mails.tsinghua.edu.cn\\
    \{miao-li, wuji\_ee\}@tsinghua.edu.cn \quad \textsuperscript{\Letter}corresponding author
}
\date{}

\begin{document}
\maketitle

\begin{abstract}
Large language models (LLMs) often struggle when low-resource training data are ambiguous or incomplete. Task-level natural-language priors can provide useful guidance in such settings, but existing approaches usually treat these priors as input context rather than as learning signals during training. We study how task-level natural-language priors can supply guidance missing from these data through Prior-Guided Tuning (PGT). Its implementation, Contrastive Prior Steering (CPS), retains the supervised objective and adds positive- and negative-prior-conditioned losses with opposing optimization directions.  Across AmbiMath, Diagnostic Anchoring, and Jigsaw, CPS achieves near-perfect exact-match accuracy on the synthetic task, improves accuracy under misleading diagnostic suggestions by 8.2 percentage points, and improves average subgroup Macro F1 on Jigsaw by 10.7 percentage points over plain fine-tuning, averaged over the two evaluated models. On Jigsaw, with one tenth of the experimental training data, CPS slightly exceeds full-data plain fine-tuning. These results show how reusable task-level language can complement limited labeled examples without requiring instance-level rationales or preference pairs.
\end{abstract}

\section{Introduction}

Machine learning models, particularly large language models (LLMs), primarily acquire knowledge by fitting patterns in data distributions. However, in real-world settings, training data are rarely comprehensive, often leading models to succumb to spurious correlations or superficial cues in data-scarce and ambiguous scenarios \citep{DBLP:conf/acl/GururanganMSLBD20, DBLP:journals/corr/abs-2401-02981}. In such settings with low-resource data, the training data alone may not reveal how the task should be solved. Task-level natural-language priors, such as domain knowledge or rule descriptions, can provide critical task-relevant guidance that is missing from the empirical training distribution.

A prevalent approach to injecting priors into LLMs is training-time prompting, which appends manually designed instructions to training examples \citep{DBLP:conf/iclr/WeiBZGYLDDL22}. However, this method may be insufficient when the prior needs to disambiguate how the task should be solved. When low-resource examples support both intended and misleading rules, prior-prefixed supervision may still fail to select the intended rule. Inference-time prompting can expose the model to the prior at test time, but once fine-tuning has reinforced an unintended rule, adding the prior only at test time may be too late to change the learned behavior. We therefore study whether task-level priors written in natural language can provide useful guidance for low-resource supervised training when the training data alone are insufficient.

This gap motivates a different use of natural-language priors: task-level natural-language priors can provide useful guidance as auxiliary learning signals for low-resource LLM training. We formalize this idea as \textbf{Prior-Guided Tuning (PGT)}, a training perspective that integrates natural-language priors as auxiliary signals to guide the training process. Within PGT, we introduce \textbf{Contrastive Prior Steering (CPS)}, a concrete implementation that combines the standard supervised loss with contrastive positive and negative prior-conditioned objectives. CPS encourages task-consistent learning while discouraging plausible but misleading alternatives, so that natural-language priors can complement sparse supervision during training. Our main contributions are as follows.

\begin{itemize}

\item  We develop \textbf{P}rior-\textbf{G}uided \textbf{T}uning (\textbf{PGT}), a training perspective that treats natural-language priors as explicit learning signals rather than contextual inputs. PGT allows models to use task-level guidance when low-resource data alone are insufficient.

\item We introduce \textbf{C}ontrastive \textbf{P}rior \textbf{S}teering (\textbf{CPS}), a concrete implementation of PGT that constructs contrastive auxiliary signals from positive and negative priors to encourage task-consistent learning and discourage misleading alternatives, while preserving the original data-driven learning signal.

\item We conduct extensive experiments on AmbiMath, Diagnostic Anchoring, and Jigsaw toxicity classification~\citep{do2019jigsaw}, covering controlled rule ambiguity, misleading diagnostic suggestions, and identity-related classification. Comparisons with prior-informed baselines, data-scaling experiments, and early-training diagnostics examine the effectiveness, data efficiency, and optimization behavior of CPS.

\end{itemize}

\section{Related Work}

Shortcut learning can yield strong in-distribution performance while relying on cues that do not generalize \citep{geirhos2020shortcut}.
In NLP, hypothesis-only label cues and syntactic heuristics illustrate how training data can reward such alternatives \citep{gururangan-etal-2018-annotation,DBLP:conf/acl/McCoyPL19}.
We study the low-resource setting in which labeled examples may not distinguish the intended rule from a misleading one, while that distinction can be stated in task-level natural language.
PGT uses this additional guidance to construct training objectives; CPS implements it through positive- and negative-prior-conditioned losses.

\subsection{Instruction Tuning and Prompting}

In-context prompting provides instructions or demonstrations \citep{DBLP:conf/nips/BrownMRSKDNSSAA20}, whereas prompt tuning learns continuous prompts \citep{DBLP:conf/emnlp/LesterAC21}.
T0 and FLAN train on instruction-formatted tasks \citep{DBLP:conf/iclr/SanhWRBSACSRDBX22,DBLP:conf/iclr/WeiBZGYLDDL22}, and Super-NaturalInstructions extends instruction tuning to broad task generalization \citep{DBLP:conf/emnlp/WangMAKMNADASPK22}.
These approaches supply guidance through input context.
PGT additionally uses task-level priors as auxiliary objectives when the task is specified but scarce examples leave its decision rule ambiguous.

\subsection{Knowledge-aware Learning}

ERNIE and K-BERT inject structured knowledge into representations or inputs \citep{DBLP:journals/corr/abs-1904-09223,DBLP:conf/aaai/LiuZ0WJD020}.
KPT expands prompt verbalizers with external knowledge, while KaFT adjusts training examples according to knowledge conflict \citep{DBLP:conf/acl/HuDWLWLWS22,zhong2025kaftknowledgeawarefinetuningboosting}.
PGT uses reusable natural-language task priors as training signals, rather than structured knowledge or additional instance-level paired annotations.

\begin{figure*}[tbp]
    \centering
    \includegraphics[width=0.86\textwidth]{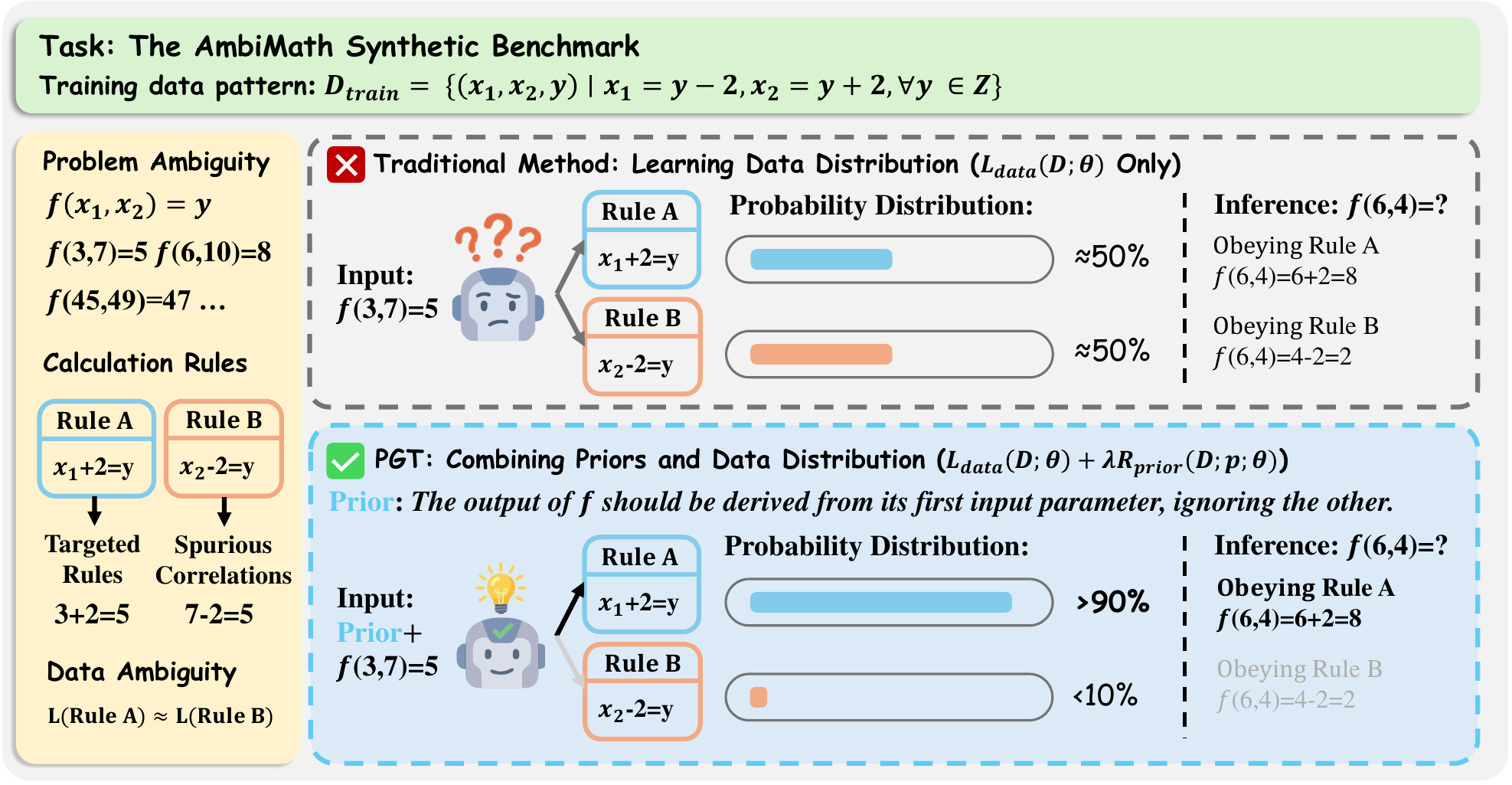}
    \caption{\textcolor{black}{Comparison between standard fine-tuning and prior-guided tuning (PGT) on AmbiMath.} Standard fine-tuning is trained on examples compatible with multiple rules. PGT uses a task-level natural-language prior as an auxiliary training signal, helping select the intended hypothesis while preserving data-driven operator learning.}
    \label{fig:PGT}
\end{figure*}

\section{Motivation: The Ambiguity of Data}

\subsection{Problem Setup}

Standard LLM training operates under the Maximum Likelihood Estimation (MLE) framework, optimizing parameters $\theta$ to maximize $P(Y|X;\theta)$ over a training dataset $\mathcal{D}$. Successful generalization depends in part on whether the training data distinguish the intended task rule from plausible alternatives. However, low-resource datasets are especially prone to specification ambiguity \citep{DBLP:journals/jmlr/DAmourHMAABCDEH22}. A small training set may lack counterexamples that distinguish the intended rule from plausible alternatives. As a result, multiple hypotheses $\{h_1,\dots,h_k\}$ can obtain nearly indistinguishable empirical losses on the training distribution but diverge on out-of-distribution samples.

Consider a scenario where a spurious feature $x_{spurious}$ strongly, or even perfectly, correlates with the label $y$ within $\mathcal{D}$. The empirical objective may fail to distinguish the intended rule from a spurious rule:
\begin{equation}
\mathcal{L}_{\mathcal{D}}(\theta_{true})
\approx
\mathcal{L}_{\mathcal{D}}(\theta_{spurious}),
\end{equation}
where $\theta_{true}$ and $\theta_{spurious}$ denote models following the intended and misleading rules, respectively. In such cases, the training data alone may not break this empirical tie. The model therefore benefits from external learning signals, such as guidance from natural-language priors, which favor the task-consistent solution over misleading alternatives like spurious correlations.

\subsection{AmbiMath: A Controlled Synthetic Benchmark}

In real-world datasets, the contributions of data distributions and guidance from natural-language priors to model performance are often deeply intertwined, making it difficult to isolate and quantify the specific impact of priors. To better separate the role of prior guidance from data-driven learning, we use a highly controlled environment. As illustrated in Figure \ref{fig:PGT}, we introduce \textbf{AmbiMath} (Ambiguous Mathematical Reasoning Benchmark), a synthetic benchmark based on function calculation tasks $f(x_1, x_2) = y$, where only one of the two input parameters determines the answer $y$, while the other is irrelevant. In this setup, the model is tasked with learning two distinct components: (1) \textbf{Feature Selection}: identifying which parameter is task-relevant, and (2) \textbf{Operator Learning}: identifying the required arithmetic operation. We construct the training set $\mathcal{D}_{train}$ as follows:
\begin{equation}
\mathcal{D}_{train} = \{(y-2, y+2, y) \mid y \in \mathbb{Z} \}
\end{equation}
Under this construction, two competing hypotheses emerge: the \textbf{Target Rule} ($h_{true}: f(x_1, x_2) = x_1 + 2$) and the \textbf{Spurious Rule} ($h_{spurious}: f(x_1, x_2) = x_2 - 2$). For any sample in $\mathcal{D}_{train}$, both rules are exactly consistent with the observed label. Thus, the symbolic data construction creates an empirical tie: the training set alone does not identify which rule should govern disentangled test cases. To resolve the task ambiguity, we provide a natural-language prior: \textit{``The output should be derived from the first input parameter, ignoring the other.''} This prior addresses the Feature Selection task, while the arithmetic operation must still be learned from the data distribution, ensuring that priors and data distributions are conceptually decoupled, and both are essential for the task. During inference, we evaluate without priors to test whether the model has already learned the target rule instead of the spurious rule. We test models on disentangled samples where $x_1+2 \neq x_2-2$ (e.g., $f(17, 5)$). If the model has effectively learned the intended rule, it should output $19$ ($17+2$); otherwise, it may output $3$ ($5-2$), indicating a failure to break the statistical tie and counteract spurious correlations.

We curated a compact dataset comprising 200 samples to simulate a low-resource scenario. To test whether CPS can use different descriptions of the same task-level rule, we vary how the relevant parameters $x_1$ and $x_2$ are expressed: the first parameter is written in Chinese numerals and the second in English words. We further evaluate two forms of priors: \textbf{positional priors} (e.g., ``focus on the first/second parameter") and \textbf{semantic priors} (e.g., ``focus on the Chinese/English parameter"). This setup tests whether the method can use different natural-language descriptions of the task-level selection rule, rather than relying only on one fixed prompt template.

\begin{figure*}[tbp]
    \centering
    \makebox[\textwidth][c]{\includegraphics[width=1.0\textwidth]{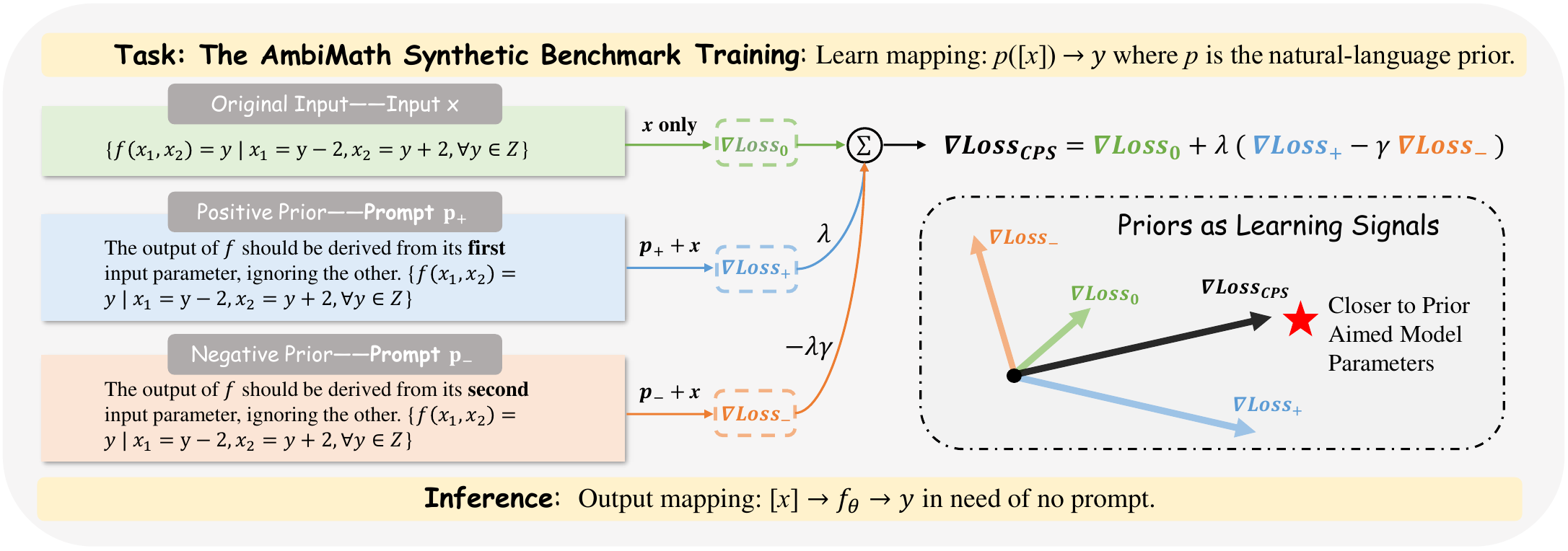}}
    \caption{\textcolor{black}{Illustration of Contrastive Prior Steering (CPS) on AmbiMath.} CPS keeps the original training view and constructs two contrastive auxiliary views, turning task-level natural-language priors into learning signals. The positive prior describes the intended task principle, while the negative prior describes a plausible but misleading alternative. Their target-token losses are combined with the original supervised loss.}\label{fig:CPS}
\end{figure*}

\section{Method}

\subsection{Prior-Guided Tuning}

Low-resource data may not distinguish the intended task rule from misleading alternatives.
Practitioners can often express useful guidance in natural language, such as which feature to trust, which heuristic to avoid, or which domain criterion to apply.
PGT uses this task-level knowledge without requiring additional instance-level annotations.

We propose \textbf{Prior-Guided Tuning} (\textbf{PGT}), a training framework that uses task-level natural-language priors as auxiliary learning signals. A prior is a task-level natural-language statement shared across examples of the same task. Unlike prompt-based fine-tuning, where such language is appended to each input as additional context, PGT uses the prior to define an auxiliary training objective. The general form is
\begin{equation}
\mathcal{L}_{\mathrm{PGT}}(\theta)
=
\mathcal{L}_{\mathrm{data}}(\mathcal{D};\theta)
+
\lambda \mathcal{R}_{\mathrm{prior}}(\mathcal{D},p;\theta),
\end{equation}
where $\mathcal{L}_{\mathrm{data}}$ is the standard supervised loss, $p$ denotes task-level natural-language priors, $\mathcal{R}_{\mathrm{prior}}$ is a prior-induced auxiliary objective, and $\lambda$ controls its strength.

PGT is intended to complement, rather than replace, data-driven learning. Figure~\ref{fig:PGT} illustrates this role: when scarce data leave multiple solutions empirically plausible, the prior supplies task-level guidance that is difficult to infer from the data alone. At the same time, the model must still learn the remaining task structure from examples. In AmbiMath, for instance, the prior specifies which input parameter is relevant, while the training examples provide the arithmetic relationship. In real tasks, analogous priors may specify invariances, domain constraints, or misleading heuristics to avoid. Thus, PGT provides a lightweight interface for incorporating task-level knowledge into low-resource training without requiring additional labeled examples for each instance.

\begin{table*}[tbp]
\centering
\fontsize{8}{10}\selectfont
\renewcommand{\arraystretch}{1.10}
\setlength{\tabcolsep}{4pt}
\begin{tabular*}{0.96\textwidth}{l@{\extracolsep{\fill}}cccc}
\toprule
\textbf{Method} & \textbf{First Param.} & \textbf{Second Param.} & \textbf{Chinese Param.} & \textbf{English Param.} \\
\midrule
Plain  & 65.3 & 34.7 & 65.3 & 34.7 \\
Prompt & 54.2 & 51.6 & 56.3 & 36.8 \\
Dual-Prior IFT & 77.9 & 77.4 & 69.0 & 84.7 \\
Multi-view SFT & 60.0 & 84.2 & 62.1 & 96.3 \\
CPS    & \textbf{96.3} & \textbf{100.0} & \textbf{96.8} & \textbf{97.4} \\
\bottomrule
\end{tabular*}
\vspace{4pt}
\caption{\textcolor{black}{Exact-match (EM) accuracy (\%) on AmbiMath using LLaMA 3.1 8B.} Columns correspond to different task-level priors specifying the relevant input parameter.}
\label{tab:syn_results}
\end{table*}

\subsection{Contrastive Prior Steering}

We instantiate PGT with \textbf{Contrastive Prior Steering} (\textbf{CPS}). CPS uses two task-level priors: a positive prior $p_{+}$ that describes task-consistent guidance, and a negative prior $p_{-}$ that describes a misleading but plausible alternative. Figure~\ref{fig:CPS} illustrates that for each labeled example $(I_i,x_i,y_i)$, where $I_i$ denotes the task instruction or prompt text, CPS constructs three views: the original example, the example conditioned on the positive prior, and the example conditioned on the negative prior. We compute cross-entropy only on the target tokens $y_i$. The original supervised loss is
\begin{equation}
\mathcal{L}_{0}
=
\ell\!\left(f_{\theta}([I_i;x_i]), y_i\right).
\end{equation}
The positive-prior and negative-prior losses are
\begin{equation}
\begin{aligned}
\mathcal{L}_{+}
&=
\ell\!\left(f_{\theta}([p_{+};I_i;x_i]), y_i\right),\\
\mathcal{L}_{-}
&=
\ell\!\left(f_{\theta}([p_{-};I_i;x_i]), y_i\right).
\end{aligned}
\end{equation}
CPS then optimizes
\begin{equation}
\mathcal{L}_{\mathrm{CPS}}
=
\mathcal{L}_{0}
+
\lambda\left(\mathcal{L}_{+}-\gamma \mathcal{L}_{-}\right),
\end{equation}
where $\lambda$ balances the prior-induced objective with the original supervised objective, and $\gamma$ controls the strength of the negative prior term. This objective has two complementary effects. The positive prior term $\mathcal{L}_{+}$ encourages the model to predict the target when the input is paired with task-consistent guidance. The negative prior term $\mathcal{L}_{-}$ discourages fitting the target under a misleading prior, which is useful when scarce data alone do not rule out that alternative. The original loss $\mathcal{L}_{0}$ anchors training to the observed task distribution, so the prior objective guides the solution principle while preserving the supervised task signal. We compute cross-entropy only on target tokens and apply global gradient-norm clipping to the combined gradient before the optimizer update.
The detailed training procedure is provided in Algorithm~\ref{alg:cps_appendix} in the Appendix.

\section{Results}

\textbf{Experiment Details.}

We evaluate CPS as an implementation of PGT on three complementary settings. First, AmbiMath tests whether task-level priors can resolve a controlled ambiguity where low-resource data are compatible with multiple solution rules. 
Second, Diagnostic Anchoring tests whether CPS improves diagnosis from synthetic clinical narratives when an early suggestion is predictive during training but misleading at evaluation.
 Third, Jigsaw tests low-resource toxicity classification in real user comments, where identity terms can be spuriously associated with toxicity labels.
Together, these settings evaluate whether task-level natural-language priors can provide useful auxiliary learning signals for low-resource LLM training across a synthetic ambiguity benchmark, diagnostic anchoring in synthetic clinical narratives, and real-world identity-related toxicity classification. 

We compare CPS with four baselines.
Plain fine-tuning uses only the original task examples, while Prompt fine-tuning prefixes the positive prior to each example and minimizes supervised cross-entropy.
Dual-Prior IFT (Dual-Prior Instruction Fine-Tuning) places both priors in a single instruction, explicitly directing the model to follow the positive principle and avoid the negative heuristic, and trains on the original target answer with standard supervised cross-entropy.
This tests whether presenting both priors and their intended relationship is sufficient.
Multi-view SFT (Multi-view Supervised Fine-Tuning) uses the original, positive-prior, and negative-prior views with the same target answer, but minimizes the supervised loss on all three views.
In the AmbiMath implementation, it retains the CPS loss weights and token masks and changes only the sign of the negative-prior loss, testing whether conventional multi-view supervision accounts for the benefit of CPS.

We use LLaMA 3.1 8B \citep{DBLP:journals/corr/abs-2407-21783} for all three tasks and additionally evaluate Qwen 2.5 7B \citep{qwen2025qwen25technicalreport} on Jigsaw.
The main tables use the same prior-free inference inputs across methods to evaluate the behavior learned during training.
Appendix Table~\ref{tab:inference_results} reports an AmbiMath diagnostic with priors retained at inference.
We compute loss only on response tokens and use model-specific chat templates consistently across methods.

\textbf{AmbiMath: Synthetic Benchmark for Rule Ambiguity.}
Table~\ref{tab:syn_results} shows that CPS reaches 96.3--100.0\% EM across positional and semantic priors and outperforms all four baselines in every condition.
Plain and Prompt obtain lower scores throughout these conditions, while Dual-Prior IFT improves over both, showing that presenting the positive and negative principles together can already help.
Dual-Prior IFT nevertheless remains below CPS, so access to both priors alone does not account for the full gain.
Multi-view SFT performs unevenly across conditions and reaches 96.3\% EM for the English-parameter prior, close to CPS, but remains substantially lower in the other three conditions.
Because the AmbiMath Multi-view comparison retains the CPS views, loss weights, and token masks, the consistently higher CPS scores support the contribution of the opposing optimization direction rather than simply adding prior-conditioned supervision.
Together, these comparisons show that CPS helps select the intended rule from ambiguous training examples and retains that behavior without priors at inference.

\textbf{Diagnostic Anchoring: Robustness to Misleading Suggestions.}
We construct a five-way diagnostic anchoring setting from MedEinst \citep{chen-etal-2026-medeinst}, whose synthetic clinical narratives are derived from DDXPlus \citep{tchango2022ddxplus}.
Anchored inputs add an early diagnostic suggestion without adding clinical evidence.
The suggestion matches the gold diagnosis in 70\% of anchored training examples, making a non-evidential cue predictive of the label.
At evaluation, Clean and Anchored inputs share the same clinical evidence and diagnosis options, but Anchored inputs include \emph{a misleading suggestion}.
The prior instructs the model to treat suggestions as hypotheses rather than evidence; evaluation omits the prior, and construction details appear in Appendix~\ref{sec:diagnostic_construction}.
Table~\ref{tab:clean_anchored_results} shows that CPS achieves the highest accuracy in both conditions: 68.2\% on Clean and 45.2\% on Anchored inputs.
Its Anchored accuracy exceeds Plain and Multi-view SFT by 8.2 and 4.3 percentage points, respectively, while its Clean accuracy also improves.
Dual-Prior IFT underperforms Plain despite receiving both priors.
The substantial remaining difference between Clean and Anchored accuracy shows that misleading suggestions remain challenging; the results support improved performance under anchoring rather than complete removal of its influence.

\begin{table}[H]
\centering
\fontsize{8}{10}\selectfont
\renewcommand{\arraystretch}{1.10}
\setlength{\tabcolsep}{5pt}
\begin{tabular*}{\columnwidth}{l@{\extracolsep{\fill}}cc}
\toprule
\textbf{Method} & \textbf{Clean} & \textbf{Anchored} \\
\midrule
Plain & 65.8 & 37.0 \\
Prompt & 64.0 & 43.7 \\
Dual-Prior IFT & 58.7 & 20.7 \\
Multi-view SFT & 68.1 & 40.9 \\
CPS & \textbf{68.2} & \textbf{45.2} \\
\bottomrule
\end{tabular*}
\vspace{4pt}
\caption{Diagnostic Anchoring accuracy (\%) under clean and anchored conditions.}
\label{tab:clean_anchored_results}
\end{table}

\setlength{\tabcolsep}{1.2mm}

\begin{table}[!tbp]
\centering
\fontsize{8}{10}\selectfont
\setlength{\tabcolsep}{5pt}
\renewcommand{\arraystretch}{1.05}
\begin{tabular*}{\columnwidth}{l@{\extracolsep{\fill}}rrrr}
\toprule
\multicolumn{5}{c}{\textit{LLaMA 3.1 8B}} \\
\midrule
Method & Acc (\%) & F1+ & F1- & Macro F1 \\
\midrule
Plain & 88.3 & 0.591 & 0.931 & 0.761 \\
Prompt & 87.8 & 0.525 & 0.930 & 0.728 \\
Dual-Prior IFT & 86.7 & 0.525 & 0.922 & 0.723 \\
Multi-view SFT & 91.4 & 0.718 & 0.948 & 0.833 \\
CPS & \textbf{92.6} & \textbf{0.773} & \textbf{0.955} & \textbf{0.864} \\
\midrule
\multicolumn{5}{c}{\textit{Qwen 2.5 7B}} \\
\midrule
Method & Acc (\%) & F1+ & F1- & Macro F1 \\
\midrule
Plain & 86.3 & 0.494 & 0.920 & 0.707 \\
Prompt & 86.1 & 0.483 & 0.919 & 0.701 \\
Dual-Prior IFT & \textbf{89.9} & 0.675 & \textbf{0.940} & 0.807 \\
Multi-view SFT & 89.2 & 0.603 & 0.938 & 0.770 \\
CPS & 89.8 & \textbf{0.695} & 0.938 & \textbf{0.817} \\
\bottomrule
\end{tabular*}
\caption{\textcolor{black}{Jigsaw results averaged equally over the four gender groups.}
Accuracy is reported as a percentage; F1+ and F1- denote F1 for toxic and non-toxic labels, respectively.}
\label{tab:jigsaw_results}
\end{table}

\begin{figure*}[tbp]
    \centering
    \includegraphics[width=0.85\textwidth]{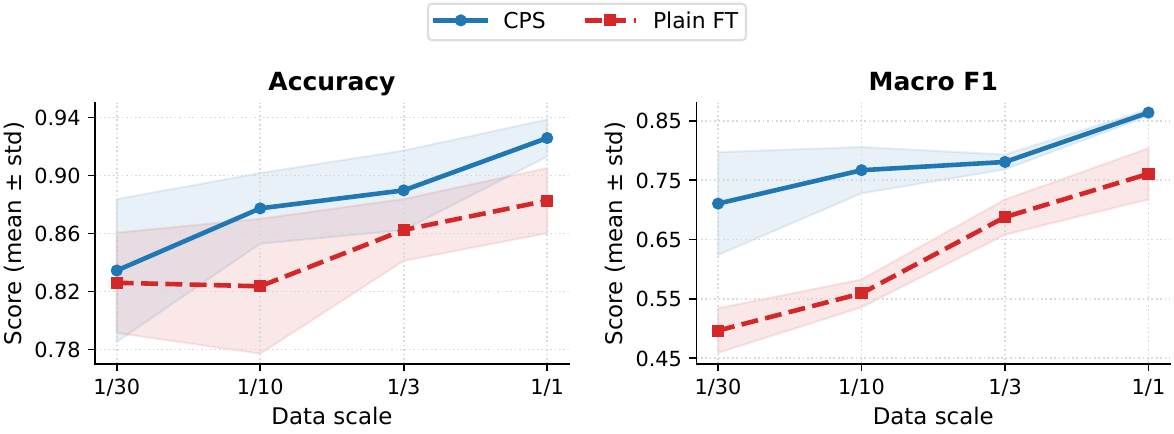}
    \caption{\textcolor{black}{Accuracy and Macro F1 of CPS and plain fine-tuning across data scales on the Jigsaw gender-associated subset.} Fractions are relative to the experimental training set used in the Jigsaw study; shaded regions denote standard deviation over three random seeds.}
    \label{fig:efficiency_scaling}
\end{figure*}

\textbf{\textcolor{black}{Jigsaw: Low-resource Toxicity Classification.}}
We evaluate CPS on gender-associated subsets of the Jigsaw toxicity dataset \citep{do2019jigsaw} using a low-resource training subset.
Table~\ref{tab:jigsaw_results} shows that Plain achieves relatively high accuracy but substantially lower F1+ than F1-, indicating weaker performance on toxic comments in these subsets.
Prompt does not improve mean subgroup Macro F1 over Plain for either model, whereas Dual-Prior IFT underperforms Plain on LLaMA but improves it on Qwen.
Multi-view SFT improves mean subgroup Macro F1 for both models, showing that ordinary supervision over the prior-conditioned views is already useful.
Dual-Prior IFT outperforms Multi-view SFT on Qwen, while Multi-view SFT is stronger on LLaMA.
CPS achieves further gains over both prior-informed baselines and the highest mean subgroup Macro F1 and F1+ for LLaMA and Qwen, improving mean subgroup Macro F1 over Plain by 10.3 and 11.0 percentage points, respectively.
On Qwen, Dual-Prior IFT achieves slightly higher accuracy and F1- than CPS, while CPS retains the best F1+ and Macro F1.
These comparisons support the value of the CPS objective for low-resource toxicity classification beyond simply presenting both priors or supervising all three views.

\section{Discussion}

The experiments above evaluate CPS as a concrete implementation of PGT. \textcolor{black}{Here we examine data efficiency, early-training gradient behavior, prior writing, and the contribution of the two prior-loss terms.}

\paragraph{Data efficiency.}
We compare CPS with plain fine-tuning on the Jigsaw gender-associated subset using 1/30, 1/10, 1/3, and all of the experimental training set.
Figure~\ref{fig:efficiency_scaling} shows gains in both accuracy and Macro F1 across data scales.
At 1/30 scale, CPS reaches roughly 0.710 Macro F1, while plain fine-tuning remains below 0.500.
With 1/10 of the data, CPS reaches 0.767 Macro F1, slightly exceeding full-data plain fine-tuning (0.761).
This result directly supports the low-resource motivation of PGT: task-level priors do not replace task data, but they can make a small amount of data substantially more useful when relevant task guidance can be stated in natural language.

\begin{figure*}[tbp]
    \centering
    \includegraphics[width=0.80\textwidth]{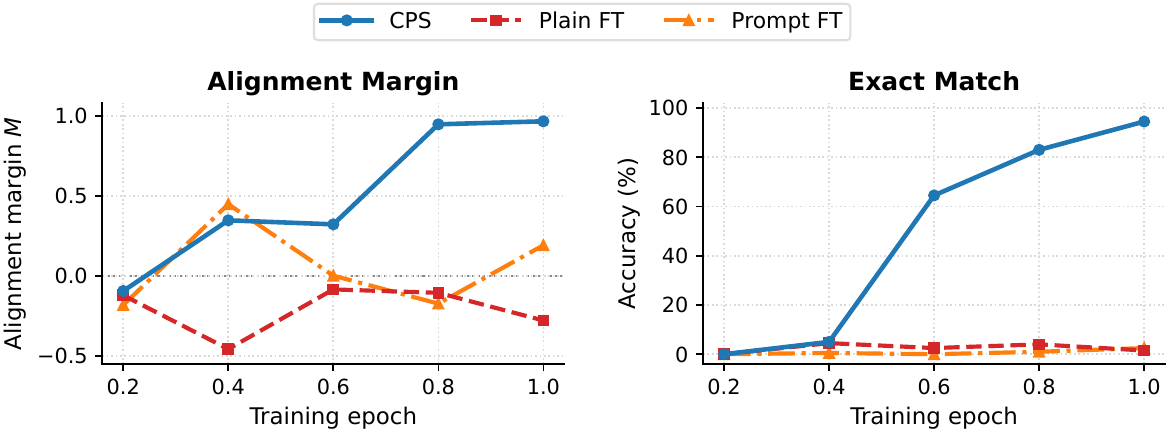}
    \caption{\textcolor{black}{Early-training diagnostic on AmbiMath.} During the first epoch, CPS rapidly improves both the alignment margin $M$ and EM accuracy, while two baseline methods remain near zero accuracy.}
    \label{fig:gradient_alignment}
\end{figure*}

\paragraph{Early-training optimization diagnostic.}
We examine early optimization on separately trained plain, prompt, and CPS trajectories in the first-parameter AmbiMath setting.
At each checkpoint, $g$ is the gradient of the corresponding training objective, averaged over all 200 training examples.
We compute supervised reference gradients $g_{\mathrm{ref}+}$ and $g_{\mathrm{ref}-}$ on the same full disentangled diagnostic set, with answers following the intended first-parameter and competing second-parameter rules, respectively.
We measure their relative alignment with $g$ as
\begin{equation}
M = \cos(g, g_{\mathrm{ref}+}) - \cos(g, g_{\mathrm{ref}-}),
\end{equation}
where larger values indicate greater alignment with the intended rule relative to the competing rule.
Figure~\ref{fig:gradient_alignment} shows that from 0.2 to 1.0 epoch, the CPS alignment margin increases from $-0.094$ to $0.966$, while EM rises from 0.0\% to 94.5\%.
At 1.0 epoch, plain fine-tuning reaches 1.5\% EM with margin $-0.278$, and prompt fine-tuning reaches 2.5\% EM with margin $0.192$.
These joint changes in gradient alignment and accuracy are consistent with priors guiding early optimization toward the intended rule, although they do not establish a complete mechanism.

\paragraph{Writing task-level priors.}
Table~\ref{tab:prior_writing} compares the original reference P0 with four alternative prior pairs: P1 states a concise insult/attack rule, P2 judges the author's stance toward offensive content, P3 focuses on identity stereotypes, and P4 focuses on quotation and endorsement.
The complete positive and negative texts are provided in Appendix~\ref{sec:prior_pair_texts}.
First, \emph{identify the task's core decision principle before elaborating details}: P1 outperforms the longer, narrowly scoped P3 and P4, suggesting that a concise rule describing what makes a comment toxic can be more useful than detailed instructions for a particular case.
P0 nevertheless performs best, so concision helps establish the core rule but should not come at the expense of useful guidance.
Second, \emph{explain how to interpret evidence, rather than merely asking the model to attend to it}: P2 asks whether the author endorses or criticizes offensive content, while P0 asks the model to judge contextual meaning rather than infer toxicity from identity words alone.
Both tell the model how to evaluate the words it encounters: determine the author's stance or whether the comment is toxic in context, rather than treating the words themselves as sufficient grounds for a label.
Third, \emph{avoid overgeneralizing misleading cues in ways that misclassify valid examples}: P0 rejects gender-word shortcuts while retaining the overall toxicity criterion, whereas P3 explicitly labels toxicity outside its selected category as non-toxic.
P0 also outperforms the more specialized P4, supporting guidance that corrects a particular error without excluding other forms of toxicity.
The resulting writing strategy is to establish the core task principle, explain how the model should interpret relevant evidence, and ensure that correcting a misleading cue does not introduce an incorrect rule for other examples.

\begin{table}[tbp]
\centering
\fontsize{8}{10}\selectfont
\setlength{\tabcolsep}{5pt}
\begin{tabular*}{\columnwidth}{l@{\extracolsep{\fill}}cc}
\toprule
Prior & Words (+/--) & Macro F1 \\
\midrule
P0 & 52/64 & 0.864 \\
P1 & 10/10 & 0.852 \\
P2 & 30/25 & 0.847 \\
P3 & 58/52 & 0.827 \\
P4 & 56/35 & 0.820 \\
\bottomrule
\end{tabular*}
\caption{Selected Jigsaw prior pairs for LLaMA 3.1 8B: mean subgroup Macro F1 (0--1). P0 is the original reference; word counts are shown for the positive and negative priors.}
\label{tab:prior_writing}
\end{table}

\begin{table}[tbp]
\centering
\fontsize{8}{10}\selectfont
\renewcommand{\arraystretch}{1.10}
\setlength{\tabcolsep}{2.5pt}
\begin{tabular*}{\columnwidth}{l@{\extracolsep{\fill}}rrrr}
\toprule
\textcolor{black}{Method} & \textcolor{black}{First} & \textcolor{black}{Second} & \textcolor{black}{Chinese} & \textcolor{black}{English} \\
\midrule
\textcolor{black}{Positive-Only CPS} & \textcolor{black}{64.7} & \textcolor{black}{91.6} & \textcolor{black}{69.5} & \textcolor{black}{73.2} \\
\textcolor{black}{Negative-Only CPS} & \textcolor{black}{65.1} & \textcolor{black}{91.8} & \textcolor{black}{26.7} & \textcolor{black}{92.1} \\
\textcolor{black}{CPS} & \textcolor{black}{\textbf{96.3}} & \textcolor{black}{\textbf{100.0}} & \textcolor{black}{\textbf{96.8}} & \textcolor{black}{\textbf{97.4}} \\
\bottomrule
\end{tabular*}
\caption{\textcolor{black}{Prior-loss component ablation on AmbiMath with LLaMA 3.1 8B: exact-match accuracy (\%).} \textcolor{black}{Columns correspond to the four priors in Table~\ref{tab:syn_results}.}}
\label{tab:component_ablation}
\end{table}

\paragraph{\textcolor{black}{Prior-loss component ablation.}}
\textcolor{black}{AmbiMath provides a controlled setting for examining the two prior-loss terms because the intended and competing rules are equally consistent with all training examples.}
\textcolor{black}{We compare full CPS with two reduced objectives, both retaining the original supervised loss: Positive-Only CPS removes $\mathcal{L}_{-}$, while Negative-Only CPS removes $\mathcal{L}_{+}$.}
\textcolor{black}{Table~\ref{tab:component_ablation} compares these variants across the four priors.}
\textcolor{black}{Both single-term variants remain below full CPS in every condition.}
\textcolor{black}{Their relative performance also depends on the prior: Positive-Only CPS is stronger for the Chinese-parameter prior, whereas Negative-Only CPS is stronger for the English-parameter prior.}
\textcolor{black}{Full CPS maintains high accuracy across all four conditions, supporting the use of both terms together in this controlled ambiguity setting.}
\textcolor{black}{Whereas Multi-view SFT tests the negative-prior update direction (Table~\ref{tab:syn_results}), this ablation tests whether both prior-conditioned terms are needed.}

\section{Conclusion}

This work studies how task-level natural-language priors can complement labeled examples when low-resource training data do not provide enough evidence to determine how a task should be solved. We formulate Prior-Guided Tuning and instantiate it with Contrastive Prior Steering, which turns positive and negative priors into auxiliary learning signals while retaining the original supervised objective. Across AmbiMath, Diagnostic Anchoring, and Jigsaw, the findings support using task knowledge not only to describe the input, but also to shape how the model learns from it.

The central lesson is that prior information and training data can serve complementary roles: examples teach the task mapping, while priors help distinguish relevant decision criteria from misleading alternatives.
The data-efficiency and prior-content analyses motivate reusable guidance that states an actionable rule without narrowing the label definition.
CPS also retains strong AmbiMath performance without priors in test prompts.

\section*{Limitations}
Our early-training gradient analysis characterizes optimization behavior but does not fully explain how prior-conditioned losses improve predictions when the priors are absent at inference.
The evaluation focuses on numeric-answer prediction, classification, and multiple-choice diagnosis; effectiveness on complex, open-ended generation remains to be established.
The present method also assumes that task-level priors are supplied before training.
Future work could use larger language models to propose, critique, and iteratively refine such priors, with validation against task-specific evidence.

\pagebreak

\clearpage
\bibliography{example_paper}
\bibliographystyle{plainnat}

\newpage
\appendix

\section{Additional Details}

\textbf{MNLI/HANS: Heuristic Robustness in NLI.}
Natural language inference (NLI) is a sentence-pair classification task: given a premise and a hypothesis, the model predicts whether the hypothesis is entailed, contradicted, or neutral with respect to the premise. In our HANS-compatible evaluation, contradiction and neutral cases are grouped as non-entailment. We fine-tune models on a low-resource MNLI subset and evaluate both in-domain MNLI accuracy and diagnostic robustness on HANS. HANS is constructed to test shallow NLI heuristics such as lexical overlap, subsequence matching, and constituent matching. Its non-entailment subset is especially important because heuristic-based models often incorrectly predict entailment on these examples. We use HANS only as a diagnostic evaluation set, not for model selection. As shown in Table~\ref{tab:hans_mnli_results}, CPS improves HANS accuracy for both LLaMA and Qwen, with gains concentrated on HANS non-entailment examples. CPS improves non-entailment accuracy from 39.2\% to 47.5\% on LLaMA and from 45.1\% to 50.3\% on Qwen, while maintaining comparable MNLI matched and mismatched accuracy. These results suggest that task-level priors can guide low-resource learning toward more robust decision principles in a different task family, where the misleading rule takes the form of shallow lexical or syntactic heuristics.

\begin{table*}[tbp]
\centering
\fontsize{8}{10}\selectfont
\renewcommand{\arraystretch}{1.10}
\setlength{\tabcolsep}{4pt}
\begin{tabular*}{0.94\textwidth}{l@{\extracolsep{\fill}}lccccc}
\toprule
\textbf{Model} & \textbf{Method} & \textbf{HANS} & \textbf{HANS Ent.} & \textbf{HANS N-Ent.} & \textbf{MNLI-m} & \textbf{MNLI-mm} \\
\midrule
\multirow{3}{*}{LLaMA 3.1 8B}
& Plain  & 69.6 & 99.9 & 39.2 & 92.0 & 93.3 \\
& Prompt & 70.6 & 98.9 & 42.3 & 91.7 & 92.4 \\
& CPS    & \textbf{73.1} & 98.7 & \textbf{47.5} & 91.5 & 93.1 \\
\midrule
\multirow{3}{*}{Qwen 2.5 7B}
& Plain  & 72.2 & 99.2 & 45.1 & 92.8 & 93.2 \\
& Prompt & 71.5 & 99.5 & 43.4 & 92.8 & 93.4 \\
& CPS    & \textbf{74.7} & 99.0 & \textbf{50.3} & 92.9 & 94.6 \\
\bottomrule
\end{tabular*}
\caption{Accuracy (\%) on MNLI and HANS. MNLI-m/mm denote matched and mismatched MNLI validation splits. HANS Ent. and HANS N-Ent. denote the entailment and non-entailment subsets of HANS; HANS N-Ent. is the key robustness metric for shallow heuristic reliance.}
\label{tab:hans_mnli_results}
\end{table*}

\paragraph{Prior formulation and component ablations.}
\textcolor{black}{The component ablations are reported in Table~\ref{tab:component_ablation}; here we examine whether CPS depends on one carefully written prior.} Table~\ref{tab:prior_writing} compares different decision rules and task coverage on Jigsaw, whereas Table~\ref{tab:ablation_results} changes the wording while retaining the same task-level meaning on AmbiMath. CPS remains substantially stronger than Plain and Prompt across these reformulations, although accuracy varies with the wording. These results support robustness to reformulations of a correct task principle, without dependence on a single prompt template. Additional appendix experiments test alternative negative-prior formulations (Table~\ref{tab:jigsaw_results_baselines_neg_prior}), prompt-retained inference (Table~\ref{tab:inference_results}), CoT-SFT diagnostics (Appendix~\ref{sec:cot_sft_diagnostics}), and operation-specifying priors (Table~\ref{tab:flexibility_results}); these diagnostics show that CPS remains effective across alternative prior formulations and is not consistently matched by prior-retained prompting or the tested generated-rationale supervision.

\setlength{\tabcolsep}{5.5pt}
\begin{table*}[tbp]
\centering
\setlength{\tabcolsep}{4pt}
\fontsize{8}{10}\selectfont
\renewcommand{\arraystretch}{1.10}
\begin{tabular}{lccccc}
\toprule
\textbf{Method} & \textbf{First Param.} & \textbf{Second Param.} & \textbf{Chinese Param.} & \textbf{English Param.} & \textbf{Avg.} \\
\midrule
Plain fine-tuning & 65.3 & 34.7 & 65.3 & 34.7 & 50.0 \\
Prompt fine-tuning & 54.2 & 51.6 & 56.3 & 36.8 & 49.7 \\
\midrule
CPS, original prior & \textbf{96.3} & \textbf{100.0} & \textbf{96.8} & \textbf{97.4} & \textbf{97.6} \\
CPS, prior variant 1 & 89.5 & \textbf{100.0} & 96.3 & 95.3 & 95.3 \\
CPS, prior variant 2 & 94.2 & 90.5 & 89.0 & 85.8 & 89.9 \\
CPS, prior variant 3 & 94.2 & 97.9 & 88.4 & 94.7 & 93.8 \\
\bottomrule
\end{tabular}
\vspace{3pt}
\caption{\textcolor{black}{Prior-formulation results on AmbiMath using LLaMA 3.1 8B.} Prior variants are manually rewritten positive and negative priors with the same task-level meaning.}
\label{tab:ablation_results}
\end{table*}

\subsection{Training Procedure and Implementation Details}

For all fine-tuning methods, we compute loss only on assistant response tokens. User instructions, system prompts, and prior text are included in the input sequence when applicable, but their tokens are masked out of the supervised loss. We use model-specific chat templates for LLaMA 3.1 and Qwen 2.5 and keep the template fixed across plain fine-tuning, prompt fine-tuning, and CPS.

For each training example $(I_i,x_i,y_i)$, CPS constructs three views: the original example, a positive-prior view, and a negative-prior view. The original supervised loss is $\mathcal{L}_{0}$, while $\mathcal{L}_{+}$ and $\mathcal{L}_{-}$ are the target-token losses under the positive and negative priors. The training objective is
\begin{equation}
\mathcal{L}_{\mathrm{CPS}}
=
\mathcal{L}_{0}
+
\lambda\left(\mathcal{L}_{+}-\gamma \mathcal{L}_{-}\right).
\end{equation}
We use global gradient-norm clipping before the optimizer update:
\begin{equation}
\tilde{\mathbf{g}}
=
\mathbf{g}
\cdot
\min\left(1,\frac{\tau}{\|\mathbf{g}\|_2}\right),
\quad
\mathbf{g}=\nabla_\theta \mathcal{L}_{\mathrm{CPS}},
\end{equation}
with $\tau=1.0$ in our experiments. This is a standard optimization stabilizer and is applied after the combined objective is differentiated.

\begin{algorithm}[H]
   \caption{Contrastive Prior Steering}
   \label{alg:cps_appendix}
\begin{algorithmic}[1]
   \STATE {\bfseries Input:} training set $\mathcal{D}$, priors $p_{+},p_{-}$, model $f_{\theta}$, learning rate $\eta$, hyperparameters $\lambda,\gamma,\tau$
   \WHILE{not converged}
   \STATE Sample a batch $(I,x,y)\sim\mathcal{D}$
   \STATE Compute $\mathcal{L}_{0}=\ell(f_{\theta}([I;x]),y)$
   \STATE Compute $\mathcal{L}_{+}=\ell(f_{\theta}([p_{+};I;x]),y)$
   \STATE Compute $\mathcal{L}_{-}=\ell(f_{\theta}([p_{-};I;x]),y)$
   \STATE $\mathcal{L}_{\mathrm{CPS}}\leftarrow \mathcal{L}_{0}+\lambda(\mathcal{L}_{+}-\gamma\mathcal{L}_{-})$
   \STATE $\mathbf{g}\leftarrow \nabla_{\theta}\mathcal{L}_{\mathrm{CPS}}$
   \STATE $\tilde{\mathbf{g}}\leftarrow \mathbf{g}\cdot \min(1,\tau/\|\mathbf{g}\|_2)$
   \STATE Update $\theta$ using the optimizer with clipped gradient $\tilde{\mathbf{g}}$
   \ENDWHILE
   \STATE \textbf{return} $\theta$
\end{algorithmic}
\end{algorithm}

Unless otherwise stated, we train LoRA adapters with AdamW. We use rank $r=8$, LoRA scaling $\alpha=16$, dropout $0.0$, and adapt the attention and MLP projection modules \texttt{q\_proj}, \texttt{k\_proj}, \texttt{v\_proj}, \texttt{o\_proj}, \texttt{gate\_proj}, \texttt{up\_proj}, and \texttt{down\_proj}. We search learning rates in $\{1{\times}10^{-4},3{\times}10^{-4},5{\times}10^{-4}\}$ and use seeds in $\{42,256,3407\}$ for all methods. For CPS, we use $\gamma=0.1$ in the main experiments and search $\lambda$ in $\{0.3,0.5,1.0\}$. The main AmbiMath results use a fixed training recipe before evaluation on the disentangled test sets; no disentangled validation examples are used to select the task rule or choose checkpoints. For MNLI/HANS, HANS is used only for diagnostic evaluation and is not used for model selection.

\subsection{Dataset and Prior Details}

\paragraph{AmbiMath.}
AmbiMath contains function-computation examples of the form $f(x_1,x_2)=y$. In the main setting, training examples satisfy $x_1=y-2$ and $x_2=y+2$, so both $x_1+2$ and $x_2-2$ are consistent with all training labels. Evaluation examples are disentangled so that $x_1+2\neq x_2-2$. Exact match is computed on the final numeric answer. The main positive and negative priors for the first-parameter setting are:
\begin{quote}
\textit{Positive prior: The output of func should be derived from its first input parameter, ignoring the other.}
\end{quote}
\begin{quote}
\textit{Negative prior: The output of func should be derived from its second input parameter, ignoring the other.}
\end{quote}
For the other AmbiMath columns, we replace ``first'' with ``second'', ``Chinese'', or ``English'' as appropriate.

\paragraph{Diagnostic anchoring construction.}\label{sec:diagnostic_construction}
Training uses 200 distinct case families, split equally between unanchored and anchored examples; the suggested diagnosis matches the label in 70\% of the anchored examples.
Development and test sets contain 200 and 500 disjoint families, respectively. The following are the complete positive and negative priors.
\begin{quote}
\textit{Positive prior: Treat an early diagnostic suggestion as a hypothesis, not as clinical evidence.}
\end{quote}
\begin{quote}
\textit{Negative prior: Treat an early diagnostic suggestion, anecdote, or patient hypothesis as strong evidence for that diagnosis, even when the patient’s clinical findings do not independently support it.}
\end{quote}

\paragraph{Jigsaw prior pairs P0--P4.}\label{sec:prior_pair_texts}
The following are the complete English positive and negative priors for the pairs in Table~\ref{tab:prior_writing}.

\textbf{P0: Gender-aware reference.}
\begin{quote}
{\textit{Positive.} You are a toxicity classification assistant. Please judge whether the following comment text is toxic based on its context and actual meaning, not merely on identity-related terms—especially gender. Prioritize catching all toxic content and minimizing any missed toxic cases. If it is toxic, output 1; if it is not toxic, output 0.}
\end{quote}
\begin{quote}
{\textit{Negative.} You are a toxicity classification assistant. Please judge whether the following comment text is toxic based solely on the literal meaning of the words, not their actual meaning or context. Especially for gender-related terms, if they appear in the text (even in a neutral or positive context), output 1 immediately; if no such terms exist, output 0 regardless of whether the content is toxic.}
\end{quote}

\textbf{P1: General insult/attack.}
\begin{quote}
{\textit{Positive.} Insulting or attacking people through language or behavior is toxic.}
\end{quote}
\begin{quote}
{\textit{Negative.} Insulting or attacking people through language or behavior is non-toxic.}
\end{quote}

\textbf{P2: Author stance toward offensive content.}
\begin{quote}
\textit{Positive.} The presence of insulting words and profanity does not by itself mean the comment is toxic; it is necessary to consider whether the speaker endorses or criticizes the insulting content.
\end{quote}
\begin{quote}
\textit{Negative.} The presence of insulting words and profanity means the comment is toxic; there is no need to consider the speaker's stance toward the insulting content.
\end{quote}

\textbf{P3: Identity stereotypes only.}
\begin{quote}
{\textit{Positive.} For this rule, consider only one issue: identity-based stereotypes or inferiority claims. Output 1 if the comment says or implies that a gender or identity group is naturally inferior, unfit, immoral, unintelligent, overly emotional, dangerous, or deserving lower status. Output 0 for neutral identity mentions and for toxicity that does not involve an identity-based stereotype or inferiority claim.}
\end{quote}
\begin{quote}
{\textit{Negative.} For this rule, ignore identity-based stereotypes and inferiority claims. If a comment says that a gender or identity group is naturally inferior, unfit, immoral, unintelligent, overly emotional, dangerous, or deserving lower status, do not treat that as toxic by itself. Only obvious profanity or direct personal insults should make the comment toxic.}
\end{quote}

\textbf{P4: Quotation/endorsement only.}
\begin{quote}
{\textit{Positive.} For this rule, consider only one issue: whether offensive content is quoted or reported rather than endorsed. Output 0 when offensive words or claims appear only inside quotation, paraphrase, news reporting, or criticism and the author does not endorse them. Output 1 when the author adopts the quoted content or uses it as their own attack.}
\end{quote}
\begin{quote}
{\textit{Negative.} For this rule, ignore quotation and reporting context. If offensive words or claims appear anywhere in the comment, treat them as the author's own toxic statement, even when they are quoted, paraphrased, reported, or criticized.}
\end{quote}

\paragraph{MNLI/HANS.}
For MNLI/HANS, we train on a low-resource subset of 1,000 MNLI training examples and evaluate on 1,200-example MNLI matched and mismatched validation subsets, as well as the full 30,000-example HANS diagnostic set. Neutral and contradiction are grouped as non-entailment to match the HANS binary setup. The positive and negative priors are:
\begin{quote}
\textit{Positive prior: Use the full meaning of the premise and hypothesis. Do not predict entailment merely because the hypothesis words appear in the premise, or because the hypothesis is a subsequence or constituent of the premise. Output entailment only when the premise logically guarantees the hypothesis; otherwise output non-entailment.}
\end{quote}
\begin{quote}
\textit{Negative prior: Use shallow lexical heuristics. If most hypothesis words appear in the premise, or the hypothesis is a subsequence or constituent of the premise, output entailment; otherwise output non-entailment.}
\end{quote}

\paragraph{DPO-style baseline on Jigsaw.}
For the Jigsaw comparison, we additionally include an instance-level label-pair DPO baseline~\citep{rafailov2023dpo}. Each training input is paired with a preferred response containing the gold label and a rejected response containing the opposite label. We train with TRL's DPOTrainer using a PEFT reference-model setup, where \texttt{ref\_model=None} lets the trainer use the base model with adapters disabled as the reference. We search the DPO $\beta$ parameter over $\{0.1,0.3,0.5\}$ and use smaller learning rates than SFT. Table~\ref{tab:jigsaw_label_pair_dpo} reports the results, averaged equally over the four gender groups as in Table~\ref{tab:jigsaw_results}. This baseline uses different responses to the same input; it does not contrast the same gold answer under positive and negative prior-conditioned inputs. The label-pair baseline has lower Macro F1 than CPS for both models, but its results assess only the tested response-pair construction, rather than the broader effectiveness of DPO-style objectives or a prior-conditioned log-ratio alternative.

\begin{table}[tbp]
\centering
\fontsize{8}{10}\selectfont
\renewcommand{\arraystretch}{1.10}
\setlength{\tabcolsep}{2.5pt}
\begin{tabular*}{\columnwidth}{l@{\extracolsep{\fill}}rrrr}
\toprule
\textbf{Model} & \textbf{Acc.} & \textbf{F1+} & \textbf{F1-} & \textbf{Macro F1} \\
\midrule
LLaMA 3.1 8B & 89.1 & 0.661 & 0.934 & 0.797 \\
Qwen 2.5 7B & 83.5 & 0.548 & 0.898 & 0.722 \\
\bottomrule
\end{tabular*}
\caption{Jigsaw label-pair DPO results averaged equally over the four gender groups. Accuracy is reported as a percentage; F1+ and F1- denote F1 for toxic and non-toxic labels, respectively. Averages are reconstructed from the originally reported subgroup scores.}
\label{tab:jigsaw_label_pair_dpo}
\end{table}

\subsection{Additional AmbiMath Prior Variants}

Table~\ref{tab:flexibility_results} reports an additional AmbiMath setting that inverts the roles of data and priors. In the main AmbiMath setting, the data reveal the arithmetic operation, while the prior specifies which input parameter should be used. In this variant, the prior instead specifies the arithmetic operation, while the model must infer from the training distribution which parameter is consistent with that operation. The training examples remain ambiguous in the same spirit as the main benchmark: the two input parameters are constructed so that both can produce the observed output under different arithmetic mappings, and disentangled test examples reveal which mapping the model has learned.

Concretely, suppose a training example is \texttt{func(13, 17)=15}. If the prior says that ``the output of \texttt{func} should be obtained by applying $+2$ to the correct input parameter,'' the prior does not tell the model whether the first or second parameter is correct. The model must still learn from the data that $13+2=15$, so the first parameter is the one consistent with the prior-specified operation. Conversely, if the prior specifies a $-2$ operation, the same example supports the second parameter because $17-2=15$. Thus, the prior supplies the operation, while the data distribution supplies the parameter choice. At test time, we evaluate on disentangled examples where applying the prior-specified operation to the two parameters yields different answers, so exact-match accuracy reflects whether the model has combined the natural-language prior with the learned data pattern.

This setting checks that CPS is not limited to priors about parameter selection. The results are consistent with the main AmbiMath results: CPS remains substantially stronger than plain and prompt fine-tuning when the natural-language prior supplies a different component of the task. The released code includes the exact data-generation script for this variant.

\setlength{\tabcolsep}{4pt}
\begin{table}[tbp]
\centering
\setlength{\tabcolsep}{4pt}
\fontsize{8}{10}\selectfont
\renewcommand{\arraystretch}{1.10}
\begin{tabular}{lcc}
\toprule
\textbf{Method} & \textbf{Plus 2} & \textbf{Minus 2} \\
\midrule
Plain fine-tuning & 65.3 & 34.7  \\
Prompt fine-tuning & 26.3 & 77.9  \\
CPS, prior variant 1 & 87.9 & 97.4   \\
CPS, prior variant 2 & 90.5 & 94.7   \\
CPS, prior variant 3 & \textbf{91.1} & 94.7   \\
CPS, prior variant 4 & 90.0 & \textbf{99.0}   \\
\bottomrule
\end{tabular}
\vspace{4pt}
\caption{AmbiMath results when the prior specifies the arithmetic operation rather than the relevant input parameter.}
\label{tab:flexibility_results}
\end{table}

\subsection{CoT-SFT Diagnostics}
\label{sec:cot_sft_diagnostics}

A natural question is whether explicit chain-of-thought (CoT) supervision can serve the same role as CPS by converting a task-level prior into instance-level rationales. We treat this as a diagnostic rather than a broad claim about all CoT methods. The comparison is useful because CoT-SFT uses the prior in a very different way: the prior is first used to generate reasoning traces, and the student model then learns from those traces. CPS instead keeps the original answer-supervised objective and uses the prior as an auxiliary learning signal during optimization.

We first test CoT-SFT on a harder AmbiMath construction where training examples satisfy $2a+5=3b-2=c$. In the first-input target setting, the held-out answer is defined by $2a+5$; the second input remains a plausible alternative because both expressions agree on the training construction but are disentangled at test time. Here, CoT rationales are generated with access to the gold answer, giving CoT-SFT a favorable supervision format. As shown in Table~\ref{tab:cot_sft_ambimath}, CoT-SFT does not recover the target mapping, while CPS learns a substantial portion of the target rule under the same data construction. This result is consistent with the Jigsaw diagnostic: explicitly asking the model to imitate generated reasoning is not necessarily a substitute for using the prior as an auxiliary learning signal, especially when the task still requires learning part of the mapping from the data distribution.

\setlength{\tabcolsep}{6pt}
\begin{table}[tbp]
\centering
\setlength{\tabcolsep}{4pt}
\fontsize{8}{10}\selectfont
\renewcommand{\arraystretch}{1.10}
\begin{adjustbox}{max width=\columnwidth}
\begin{tabular}{lcc}
\toprule
\textbf{Method} & \textbf{Supervision format} & \textbf{Target-rule EM} \\
\midrule
Plain fine-tuning & answer only & 10.5 \\
CoT-SFT & generated rationale + answer & 0.0 \\
CPS & answer + task-level priors & \textbf{61.6} \\
\bottomrule
\end{tabular}
\end{adjustbox}
\vspace{4pt}
\caption{Diagnostic on a harder AmbiMath first-input target setting with training rule $2a+5=3b-2=c$. EM is computed against the intended first-input rule $2a+5$ on disentangled test examples.}
\label{tab:cot_sft_ambimath}
\end{table}

We also evaluate CoT-SFT on the same Jigsaw setting used in the main experiments. In the no-label setting, a strong instruction-following LLM receives the task-level prior and raw comment but not the gold toxicity label; it must generate both a rationale and a final label. Even with up to three attempts, only 79.9\% of training examples yield a correct rationale/label pair. We then train on the filtered correct CoT data. We also evaluate a more favorable answer-revealed setting, where the LLM is shown the gold label while generating the rationale. As shown in Table~\ref{tab:cot_sft_jigsaw}, both CoT-SFT variants underperform CPS by a large margin. This suggests that, when only task-level guidance is available, forcing the prior through generated rationales can introduce a brittle intermediate supervision channel: erroneous rationales either reduce the usable training set through filtering or remain as noisy reasoning targets, while the final model still does not receive the prior as a live training signal.

\setlength{\tabcolsep}{6pt}
\begin{table}[tbp]
\centering
\setlength{\tabcolsep}{4pt}
\fontsize{8}{10}\selectfont
\renewcommand{\arraystretch}{1.10}
\begin{adjustbox}{max width=\columnwidth}
\begin{tabular}{lccc}
\toprule
\textbf{Method} & \textbf{Accuracy} & \textbf{Macro F1} & \textbf{Toxic F1} \\
\midrule
CoT-SFT, no gold label (filtered) & 82.3 & 0.700 & 0.508 \\
CoT-SFT, gold label revealed & 83.6 & 0.690 & 0.478 \\
CPS & \textbf{92.6} & \textbf{0.864} & \textbf{0.773} \\
\bottomrule
\end{tabular}
\end{adjustbox}
\vspace{4pt}
\caption{CoT-SFT diagnostic on the Jigsaw toxicity setting. Accuracy is reported as a percentage; F1 scores are reported on a 0--1 scale. ``No gold label'' means the rationale generator sees only the task-level prior and raw comment; ``gold label revealed'' gives the generator the answer when writing the rationale. The CPS row averages the four gender categories reported in Table~\ref{tab:jigsaw_results} for LLaMA 3.1 8B.}
\label{tab:cot_sft_jigsaw}
\end{table}

\subsection{Bounded Surrogate Diagnostic}

CPS uses the negative prior by reducing the likelihood of the gold answer under a misleading prior-conditioned view. This makes the auxiliary contrastive term a direct training signal rather than a bounded surrogate. In our intended low-resource setting, we keep this objective controlled by retaining the original supervised loss, using a small negative-prior coefficient $\gamma=0.1$, and applying global gradient-norm clipping. Across the reported runs, we did not observe numerical divergence.

We also tested whether a bounded surrogate could replace the direct negative-prior term. Specifically, we trained a variant of the form
\begin{equation}
\mathcal{L}_{\mathrm{bnd}}
=
\mathcal{L}_{0}
+
\lambda_b\,\sigma\!\left(\mathcal{L}_{+}-\gamma_b\mathcal{L}_{-}\right),
\end{equation}
where $\sigma(\cdot)$ is the sigmoid function. This keeps the auxiliary term bounded, but it can also saturate and provide a weaker corrective signal after the model begins separating the positive and negative prior views. Table~\ref{tab:bounded_variant} reports the best bounded run from our AmbiMath sweep. The bounded variant improves some settings but is much less consistent than CPS, especially on the first-parameter and Chinese-prior conditions. We therefore use the direct CPS objective in the main experiments, while viewing better bounded formulations as a useful direction for future work.

\setlength{\tabcolsep}{5pt}
\begin{table*}[tbp]
\centering
\setlength{\tabcolsep}{4pt}
\fontsize{8}{10}\selectfont
\renewcommand{\arraystretch}{1.10}
\begin{tabular}{lccccc}
\toprule
\textbf{Method} & \textbf{First} & \textbf{Second} & \textbf{Chinese} & \textbf{English} & \textbf{Avg.} \\
\midrule
Bounded surrogate & 54.7 & 95.3 & 20.0 & 87.9 & 64.5 \\
CPS & \textbf{96.3} & \textbf{100.0} & \textbf{96.8} & \textbf{97.4} & \textbf{97.6} \\
\bottomrule
\end{tabular}
\vspace{4pt}
\caption{AmbiMath EM comparison between CPS and the best bounded surrogate run. The bounded surrogate avoids direct negative-loss subtraction but provides weaker and less consistent prior guidance.}
\label{tab:bounded_variant}
\end{table*}

\begin{table*}[tbp]
\centering
\fontsize{8}{10}\selectfont
\setlength{\tabcolsep}{4pt}
\renewcommand{\arraystretch}{1.10}
\begin{adjustbox}{max width=\textwidth}
\begin{tabular}{lccccc}
\toprule
\textbf{Method} & \textbf{Scale} & \textbf{Steps} & \textbf{Train time} & \textbf{Time/step} & \textbf{Peak memory} \\
\midrule
Plain FT & 1/3  & 1616 & 1303s & 0.81s & 11.53 GiB \\
CPS      & 1/30 & 162  & 1090s & 6.73s & 11.53 GiB \\
\bottomrule
\end{tabular}
\end{adjustbox}
\vspace{4pt}
\caption{Training cost profile on the Jigsaw data-efficiency setting. CPS is more expensive per step but uses fewer examples and optimization steps in this cost-to-target comparison.}
\label{tab:compute_cost}
\end{table*}

\subsection{Robustness to Negative Prior Formulation}

In this Jigsaw diagnostic, the positive prior is held fixed while the negative prior is varied. The default negative prior, Neg-Prior1, describes the direct identity-term shortcut: mention of a gender or identity group is treated as the main toxicity signal. We also test two alternative ways of constructing a misleading negative signal. Neg-Prior2 replaces the direct shortcut with a vague subjective judgment based on rough associations with gender terms. Neg-Prior3 instead encourages under-detection by asking the model to ignore toxic content unless the toxicity is explicit. The concrete negative priors are:
\begin{quote}
\textit{Neg-Prior1 (default): Use identity-related terms as the main signal. If the comment mentions a gender or identity group, predict toxic; otherwise predict non-toxic.}
\end{quote}
\begin{quote}
\textit{Neg-Prior2: Judge toxicity based on a subjective impression. If the text roughly feels associated with gender-related groups, treat it as toxic; otherwise treat it as non-toxic.}
\end{quote}
\begin{quote}
\textit{Neg-Prior3: Be conservative about toxicity detection. Unless the comment contains explicit toxic language, ignore potentially toxic content and predict non-toxic.}
\end{quote}
As shown in Table~\ref{tab:jigsaw_results_baselines_neg_prior}, both alternative formulations retain strong performance, suggesting that CPS is not tied to a single handcrafted negative-prior wording.

\setlength{\tabcolsep}{3.5pt}
\begin{table}[H]
\centering
\setlength{\tabcolsep}{4pt}
\fontsize{8}{10}\selectfont
\renewcommand{\arraystretch}{1.10}
\begin{tabular}{llcccc}
\toprule
\textbf{Method} & \textbf{Gender} & \textbf{Acc} & \textbf{F1+} & \textbf{F1-} & \textbf{Macro F1} \\
\midrule
\multirow{4}{*}{Neg-Prior2}
& female & 94.1 & 0.764 & 0.966 & 0.865 \\
& male & 94.0 & 0.780 & 0.965 & 0.873 \\
& other & 91.2 & 0.800 & 0.943 & 0.872 \\
& trans & 90.4 & 0.714 & 0.943 & 0.828 \\
\midrule
\multirow{4}{*}{Neg-Prior3}
& female & 93.7 & 0.747 & 0.964 & 0.856 \\
& male & 92.8 & 0.741 & 0.958 & 0.850 \\
& other & 94.1 & 0.875 & 0.962 & 0.918 \\
& trans & 89.0 & 0.676 & 0.934 & 0.805 \\
\bottomrule
\end{tabular}
\vspace{4pt}
\caption{Jigsaw results with alternative negative priors for LLaMA 3.1 8B. F1+ and F1- denote toxic and non-toxic labels.}
\label{tab:jigsaw_results_baselines_neg_prior}
\end{table}

\subsection{Sensitivity to the Prior-Loss Weight}

We sweep the CPS prior-loss weight with LLaMA 3.1 8B on the 1/30-data Jigsaw setting, the smallest training scale in Figure~\ref{fig:efficiency_scaling}, while keeping other hyperparameters fixed. Results are from seed 42 and are averaged equally over the four gender groups.
This sweep concerns the minimum-data regime rather than the main Jigsaw comparison in Table~\ref{tab:jigsaw_results}.
Table~\ref{tab:lambda_sensitivity} shows that CPS has a useful operating range rather than a single narrow setting: $\lambda=0.3$ and $\lambda=0.5$ are nearly identical, and $\lambda=1.0$ remains close. Very small or overly weak prior weighting is less effective, so we do not claim that CPS is hyperparameter-free; the result instead suggests that the method does not rely on a fragile, cherry-picked value.

\setlength{\tabcolsep}{7pt}
\begin{table}[H]
\centering
\setlength{\tabcolsep}{4pt}
\fontsize{8}{10}\selectfont
\renewcommand{\arraystretch}{1.10}
\begin{tabular}{ccc}
\toprule
\textbf{$\lambda$} & \textbf{Avg. Macro F1} & \textbf{Avg. F1+} \\
\midrule
0.1 & 0.646 & 0.389 \\
0.2 & 0.699 & 0.490 \\
0.3 & 0.737 & 0.560 \\
0.4 & 0.710 & 0.521 \\
0.5 & \textbf{0.738} & \textbf{0.562} \\
0.6 & 0.705 & 0.506 \\
0.8 & 0.664 & 0.417 \\
1.0 & 0.733 & 0.551 \\
\bottomrule
\end{tabular}
\vspace{4pt}
\caption{Sensitivity to the CPS prior-loss weight on Jigsaw with LLaMA 3.1 8B at the 1/30 training-data scale used in Figure~\ref{fig:efficiency_scaling}.}
\label{tab:lambda_sensitivity}
\end{table}

\subsection{Inference with and without Priors}

Table~\ref{tab:inference_results} compares AmbiMath evaluation with and without priors at inference time. CPS remains strong when the prior is removed, while prompt fine-tuning benefits more from reintroducing the prior at inference. This supports the main paper's distinction between using priors as training signals and using them only as input context. For CPS, retaining the prior improves the first-parameter result but lowers the other three, showing that its training-time benefit does not imply a uniform inference-time benefit. We present this result as a diagnostic rather than as the main contribution.

\setlength{\tabcolsep}{4pt}
\begin{table}[H]
\centering
\setlength{\tabcolsep}{4pt}
\fontsize{8}{10}\selectfont
\renewcommand{\arraystretch}{1.10}
\begin{tabular}{lcccc}
\toprule
\textbf{Method} & \textbf{First} & \textbf{Second} & \textbf{Chinese} & \textbf{English} \\
\midrule
\multicolumn{5}{l}{\textit{Inference without priors}} \\
Plain fine-tuning & 65.3 & 34.7 & 65.3 & 34.7 \\
Prompt fine-tuning & 54.2 & 51.6 & 56.3 & 36.8 \\
CPS & 96.3 & \textbf{100.0} & \textbf{96.8} & \textbf{97.4} \\
\midrule
\multicolumn{5}{l}{\textit{Inference with priors}} \\
Plain fine-tuning & 69.5 & 36.8 & 52.1 & 34.7 \\
Prompt fine-tuning & 76.0 & 81.1 & 57.2 & 76.3 \\
CPS & \textbf{99.5} & 99.5 & 86.3 & 94.7 \\
\bottomrule
\end{tabular}
\vspace{4pt}
\caption{AmbiMath exact-match accuracy with and without priors during inference for LLaMA 3.1 8B.}
\label{tab:inference_results}
\end{table}

\subsection{Compute Profile on Jigsaw}

We profile the Jigsaw settings used for the data-efficiency comparison with LLaMA 3.1 8B. Plain fine-tuning at the 1/3 scale uses 6464 examples and 1616 optimization steps. CPS at the 1/30 scale uses 646 examples and 162 optimization steps. CPS is slower per step because it evaluates the original, positive-prior, and negative-prior views and the prior-conditioned views are longer. The relevant comparison here is therefore cost-to-target in the low-resource regime, not per-step cost. In this profile, CPS reaches a higher Macro F1 than the larger plain fine-tuning run with fewer optimization steps and lower measured wall-clock training time.

\subsection{Ethics and Broader Impacts}

This work studies how to use task-level natural-language priors during low-resource LLM training. The main intended benefit is to reduce reliance on misleading correlations when task knowledge is available. The same mechanism can also reflect errors in the supplied priors, so priors should be reviewed before deployment, especially in high-stakes applications. The Jigsaw experiments involve toxic or identity-related language from a public benchmark; such content is used only for evaluation and does not express the authors' views. We do not collect private data or conduct new human-subject studies.

\subsection{Reproducibility Statement}

We will release code and data needed to reproduce the main experiments, including AmbiMath generation scripts, training and evaluation scripts, prior prompts, model identifiers, and hyperparameter configurations. All datasets used in the paper are public or generated synthetically. Experiments were conducted by fine-tuning pretrained language models on standard GPU servers; model-specific chat templates and assistant-only loss masking are used consistently across methods.

\subsection{Use of Large Language Models}

The authors used LLM-based tools to assist with grammar checking, wording polish, and literature search during paper preparation. The research questions, experimental design, implementation, result interpretation, and final text were reviewed and controlled by the authors. LLM tools were not used to generate experimental results or make autonomous scientific decisions.

\end{document}